\documentclass[10pt,twocolumn,letterpaper]{article}
    
\usepackage{color}

\usepackage{cvpr}
\usepackage{url}
\usepackage[T1]{fontenc}
\usepackage{newtxtext,newtxmath}
\usepackage{graphicx}
\usepackage{amsmath}
\usepackage{amssymb}
\usepackage{booktabs}
\usepackage{multirow}
\usepackage{subcaption}
\usepackage{caption}
\usepackage{array}
\newcolumntype{M}[1]{>{\centering\arraybackslash}m{#1}}
\newcolumntype{L}[1]{>{\arraybackslash}m{#1}}

\newcommand{\oneone}[2]{
  \(
    \begin{array}{l}
      \{ #1, \\
      \hspace{1.7mm} #2 \} \\
    \end{array}
  \)
}
\newcommand{\twotwo}[4]{
  \(
    \begin{array}{ll}
      \{ #1, #2, \\
      \hspace{1.7mm} #3, #4 \} \\
    \end{array}
  \)
}

\usepackage[pagebackref=true,breaklinks=true,letterpaper=true,colorlinks,bookmarks=false]{hyperref}

\cvprfinalcopy

\def\cvprPaperID{482} 

\ifcvprfinal\fi
\begin{document}

\title{Can Spatiotemporal 3D CNNs Retrace the History of 2D CNNs and ImageNet?}

\author{Kensho Hara, Hirokatsu Kataoka, Yutaka Satoh\\
National Institute of Advanced Industrial Science and Technology (AIST)\\
Tsukuba, Ibaraki, Japan\\
{\ttfamily\small \{kensho.hara, hirokatsu.kataoka, yu.satou\}@aist.go.jp}
}

\maketitle

\begin{abstract}
  The purpose of this study is to determine whether current video datasets have sufficient data for
  training very deep convolutional neural networks (CNNs) with spatio-temporal three-dimensional (3D) kernels.
  Recently, the performance levels of 3D CNNs in the field of action recognition have improved significantly.
  However, to date, conventional research has only explored relatively shallow 3D architectures.
  We examine the architectures of various 3D CNNs from relatively shallow to very deep ones on current video datasets.
  Based on the results of those experiments, the following conclusions could be obtained:
  (i) ResNet-18 training resulted in significant overfitting for UCF-101, HMDB-51, and ActivityNet but not for Kinetics.
  (ii) The Kinetics dataset has sufficient data for training of deep 3D CNNs,
  and enables training of up to 152 ResNets layers, interestingly similar to 2D ResNets on ImageNet.
  ResNeXt-101 achieved 78.4\% average accuracy on the Kinetics test set.
  (iii) Kinetics pretrained simple 3D architectures outperforms complex 2D architectures,
  and the pretrained ResNeXt-101 achieved 94.5\% and 70.2\% on UCF-101 and HMDB-51, respectively.
  
  The use of 2D CNNs trained on ImageNet has produced significant progress in various tasks in image.
  We believe that using deep 3D CNNs together with Kinetics will retrace the successful history of 2D CNNs and ImageNet,
  and stimulate advances in computer vision for videos.
  The codes and pretrained models used in this study are publicly available\footnote{\url{https://github.com/kenshohara/3D-ResNets-PyTorch}}.
\end{abstract}

\section{Introduction}
  \begin{figure}[t]
    \centering
    \includegraphics[width=\linewidth, clip]{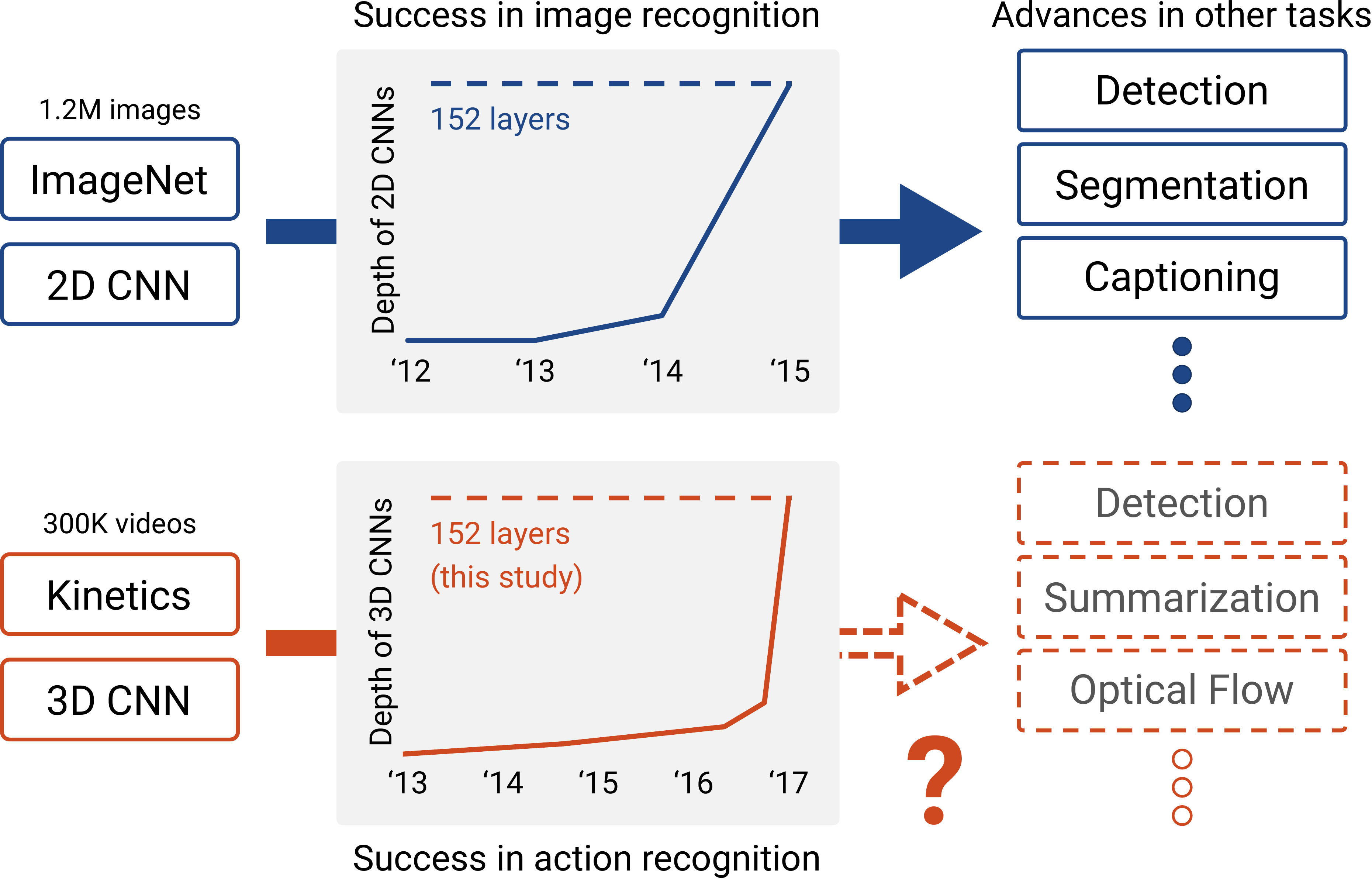}
    \caption{
      Recent advances in computer vision for images (top) and videos (bottom).
      The use of very deep 2D CNNs trained on ImageNet 
      generates outstanding progress in image recognition as well as in various other tasks.
      Can the use of 3D CNNs trained on Kinetics generates similar progress in computer vision for videos?
    }\label{fig:intro}
  \end{figure}
  The use of large-scale datasets is extremely important when using deep convolutional neural networks (CNNs),
  which have massive parameter numbers,
  and the use of CNNs in the field of computer vision has expanded significantly in recent years.
  ImageNet~\cite{imagenet_cvpr09}, which includes more than a million images,
  has contributed substantially to the creation of successful vision-based algorithms.
  In addition to such large-scale datasets, a large number of algorithms, such as residual learning~\cite{ResNet},
  have been used to improve image classification performance by adding increased depth to CNNs,
  and the use of very deep CNNs trained on ImageNet have facilitated the acquisition of generic feature representation.
  Using such feature representation, in turn, has significantly improved the performance of several other tasks 
  including object detection, semantic segmentation, and image captioning (see top row in Figure~\ref{fig:intro}).

  To date, the video datasets available for action recognition have been relatively small when compared with image recognition datasets. 
  Representative video datasets, such as UCF-101~\cite{UCF101} and HMDB-51~\cite{HMDB51},
  can be used to provide realistic videos with sizes around 10 K, 
  but even though they are still used as standard benchmarks, 
  such datasets are obviously too small to be used for optimizing CNN representations from scratch.
  In the last couple of years, ActivityNet~\cite{activitynet}, which is a somewhat larger video dataset, has become available,
  and its use has make it possible to accomplish additional tasks such as untrimmed action classification and detection,
  but the number of action instances it contains is still limited.
  More recently, the Kinetics dataset~\cite{Kinetics} was created with the aim of being positioned as
  a de facto video dataset standard that is roughly equivalent to the position held by ImageNet in relation to image datasets.
  More than 300 K videos have been collected for the Kinetics dataset, 
  which means that the scale of video datasets has begun to approach that of image datasets.

  For action recognition, CNNs with spatio-temporal three-dimensional (3D) convolutional kernels (3D CNNs)
  are recently more effective than CNNs with two-dimensional (2D) kernels~\cite{I3D}.
  From several years ago~\cite{Ji2013},
  3D CNNs are explored to provide an effective tool for accurate action recognition.
  However, even the usage of well-organized models~\cite{C3D,LongTermTemporalConv} has 
  failed to overcome the advantages of 2D-based CNNs that combine both stacked flow and RGB images~\cite{Simonyan2014}.
  The primary reason for this failure has been the relatively small data-scale of video datasets 
  that are available for optimizing the immense number of parameters in 3D CNNs, which are much larger than those of 2D CNNs.
  In addition, basically, 3D CNNs can only be trained on video datasets
  whereas 2D CNNs can be pretrained on ImageNet.
  Recently, however, Carreira and Zisserman achieved a significant breakthrough using the Kinetics dataset
  as well as the inflation of 2D kernels pretrained on ImageNet into 3D ones~\cite{I3D}.
  Thus, we now have the benefit of a sophisticated 3D convolution that can be engaged by the Kinetics dataset.
  
  However, can 3D CNNs retrace the successful history of 2D CNNs and ImageNet?
  More specifically, can the use of 3D CNNs trained on Kinetics produces significant progress in action recognition and other various tasks?
  (See bottom row in Figure~\ref{fig:intro}.)
  To achieve such progress, we consider that Kinetics for 3D CNNs should be as large-scale as ImageNet for 2D CNNs,
  though no previous work has examined enough about the scale of Kinetics.
  Conventional 3D CNN architectures trained on Kinetics are still relatively shallow
  (10~\cite{Kinetics}, 22~\cite{I3D}, and 34~\cite{Hara_2017_ICCV_Workshops,res3d} layers).
  If using the Kinetics dataset enables very deep 3D CNNs at a level similar to ImageNet, which can train 152-layer 2D CNNs~\cite{ResNet},
  that question could be answered in the affirmative.

  \begin{figure}[t]
    \centering
    \includegraphics[width=\linewidth, clip]{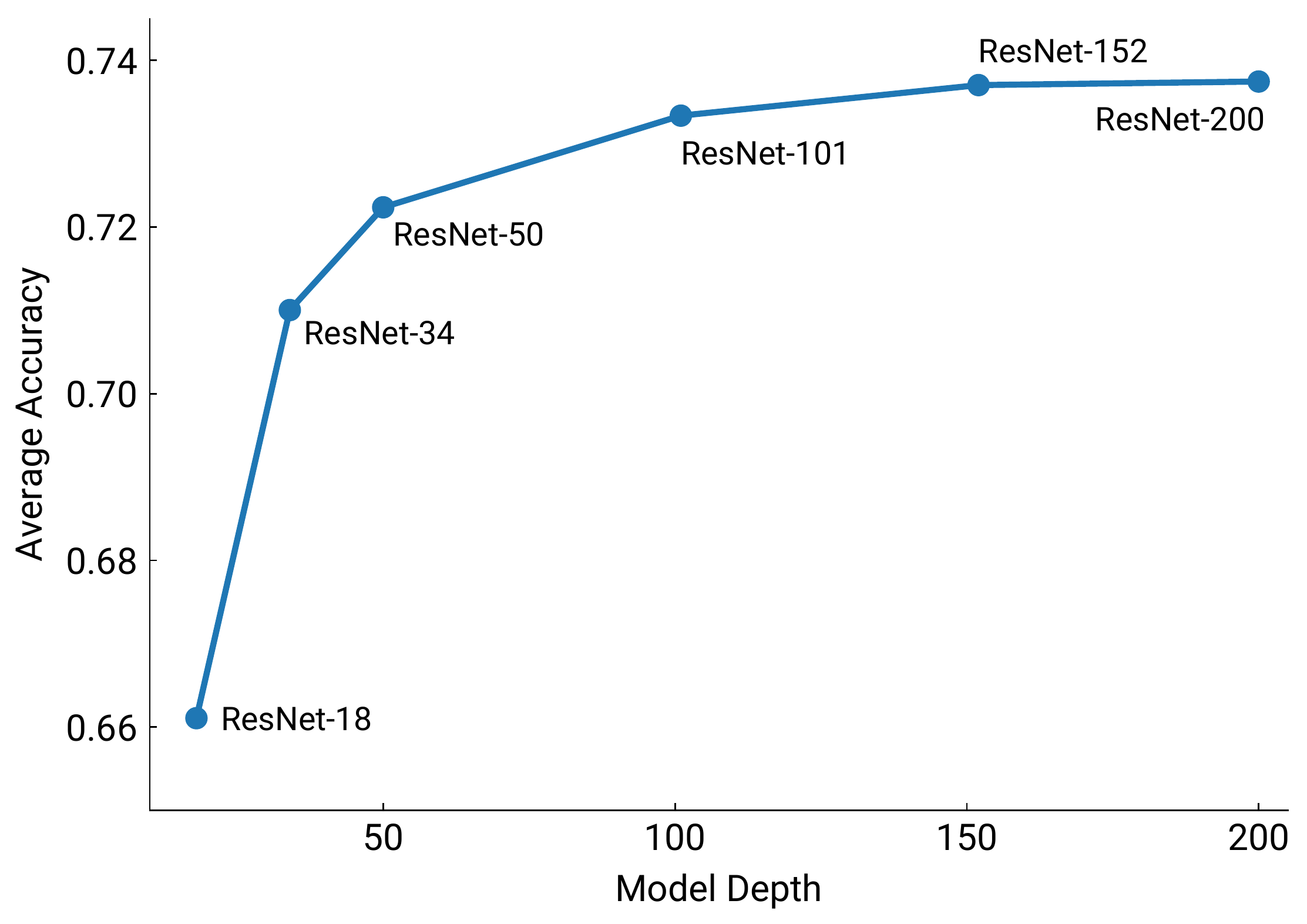}
    \caption{
      Averaged accuracies of 3D ResNets over top-1 and top-5 on the Kinetics validation set.
      Accuracy levels improve as network depths increase.
      The improvements continued until reaching the depth of 152.
      The accuracy of ResNet-200 is almost the same as that of ResNet-152.
      These results are similar to 2D ResNets on ImageNet~\cite{ResNet}.
    }\label{fig:resnet_depth}
  \end{figure}
  In this study, we examine various 3D CNN architectures from relatively shallow to very deep ones
  using the Kinetics and other popular video datasets (UCF-101, HMDB-51, and ActivityNet) in order to provide us insights for answering the above question.
  The 3D CNN architectures tested in this study are based on residual networks (ResNets)~\cite{ResNet} and their extended versions~\cite{He2016,densenets,resnext,WideResNet}
  because they have simple and effective structures.
  Accordingly, using those datasets, we performed several experiments
  aimed at training and testing those architectures from scratch, as well as their fine-tuning.
  The results of those experiments (see Section~\ref{sec:exp} for details) show the Kinetics dataset can train 3D ResNet-152 from scratch
  to a level that is similar to the training accomplished by 2D ResNets on ImageNet, as shown in Figure~\ref{fig:resnet_depth}.
  Based on those results, we will discuss the possibilities of future progress in action recognition and other video tasks.
  
  To our best knowledge, this is the first work to focus on the training of very deep 3D CNNs from scratch for action recognition.
  Previous studies showed deeper 2D CNNs trained on ImageNet achieved better performance~\cite{ResNet}.
  However, it is not trivial to show deeper 3D CNNs are better based on the previous studies
  because the data-scale of image datasets differs from that of video ones.
  The results of this study, which indicate deeper 3D CNNs are more effective,
  can be expected to facilitate further progress in computer vision for videos.

\section{Related Work}
  \subsection{Video Datasets}
    The HMDB-51~\cite{HMDB51} and UCF-101~\cite{UCF101} datasets are currently the most successful in the field of action recognition.
    These datasets gained significant popularity in the early years of the field, and are still used as popular benchmarks.
    However, a recent consensus has emerged that indicates that they are simply not large enough for training deep CNNs from scratch~\cite{Kinetics}.

    A couple of years after the abovementioned datasets were introduced, larger video datasets were produced.
    These include ActivityNet~\cite{activitynet}, which contains 849 hours of video, including 28,000 action instances.
    ActivityNet also provides some additional tasks, such as untrimmed classification and detection,
    but the number of action instances is still on the order of tens of thousands.
    This year (2017), in an effort to create a successful pretrained model,
    Kay et al.\ released the Kinetics dataset~\cite{Kinetics}.
    The Kinetics dataset includes more than 300,000 trimmed videos covering 400 categories.
    In order to determine whether it can train deeper 3D CNNs,
    we performed a number of experiments using these recent datasets, as well as the UCF-101 and HMDB-51 datasets.

    Other huge datasets such as Sports-1M~\cite{KarpathyCVPR14} and YouTube-8M~\cite{YouTube8M} have been proposed.
    Although these databases are larger than Kinetics,
    their annotations are slightly noisy and only video-level labels have been assigned.
    (In other words, they include frames that do not relate to target actions.)
    Such noise and the presence of unrelated frames have the potential to prevent these models from providing good training.
    In addition, with file sizes in excess of 10 TB, their scales are simply too large to allow them to be utilized easily.
    Because of these issues, we will refrain from discussing these datasets in this study.

  \subsection{Action Recognition Approaches}
    One of the popular approaches to CNN-based action recognition is the use of two-stream CNNs with 2D convolutional kernels.
    In their study, Simonyan et al.\ proposed a method that
    uses RGB and stacked optical flow frames as appearance and motion information, respectively~\cite{Simonyan2014},
    and showed that combining the two-streams has the ability to improve action recognition accuracy.
    Since that study, numerous methods based on the two-stream CNNs have been proposed
    to improve action recognition performance~\cite{STResNet,feichtenhofer2017multiplier,Feichtenhofer16,Wang2015TDD,VeryDeepTwo,Wang2016TSN}.
    
    Unlike the abovementioned approaches,
    we focused on CNNs with 3D convolutional kernels,
    which have recently begun to outperform 2D CNNs through the use of large-scale video datasets.
    These 3D CNNs are intuitively effective
    because such 3D convolution can be used to directly extract spatio-temporal features from raw videos.
    For example, Ji et al.\ proposed applying 3D convolution to extract spatio-temporal features from videos,
    while Tran et al.\ trained 3D CNNs, which they referred to as C3D, using the Sports-1M dataset~\cite{KarpathyCVPR14}.
    Since that study, C3D has been seen as a de facto standard for 3D CNNs.
    They also experimentally found that a \(3 \times 3 \times 3\) convolutional kernel achieved the best performance level.
    In another study, Varol et al.\ showed that expanding the temporal length of inputs for C3D improves recognition performance~\cite{LongTermTemporalConv}.
    Those authors also found that using optical flows as inputs to 3D CNNs resulted in a higher level of performance than can be obtained from RGB inputs,
    but that the best performance could be achieved by combining RGB and optical flows.
    Meanwhile, Kay et al.\ showed that the results of 3D CNNs trained from scratch on their Kinetics dataset were comparable with
    the results of 2D CNNs pretrained on ImageNet,
    even though the results of 3D CNNs trained on the UCF101 and HMDB51 datasets were inferior to the 2D CNNs results.
    In still another study, Carreira et al.\ proposed inception~\cite{Inception} based 3D CNNs, which they referred to as I3D,
    and achieved state-of-the-art performance~\cite{I3D}.
    More recently, some works introduced ResNet architectures into 3D CNNs~\cite{Hara_2017_ICCV_Workshops,res3d},
    though they examined only relatively shallow ones.

\section{Experimental configuration}
  \begin{figure*}[t]
    \centering
    \includegraphics[width=0.9\linewidth, clip]{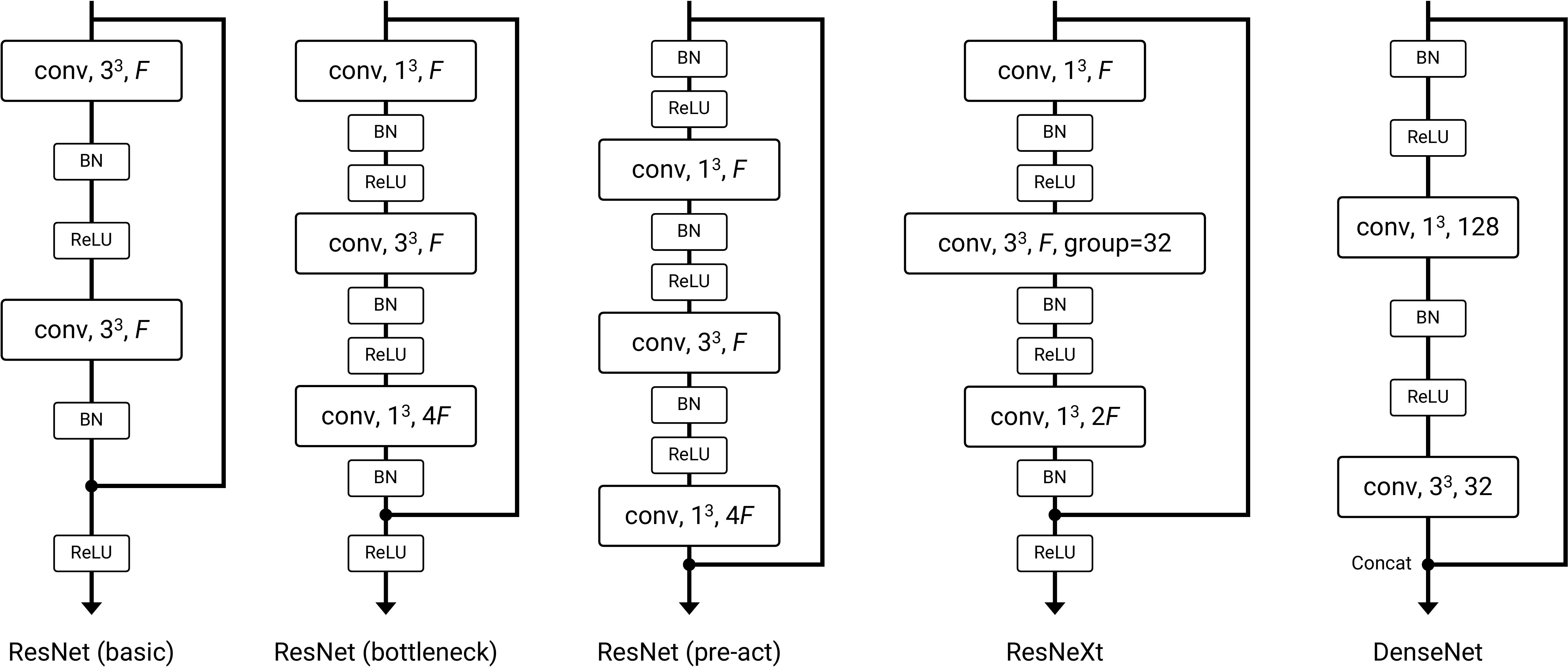}
    \caption{
      Block of each architecture.
      We represent \textit{conv}, \(x^3, F\) as the kernel size, and the number of feature maps of the convolutional filter are \(x \times x \times x\) and \(F\), respectively,
      and \textit{group} as the number of groups of group convolutions, which divide the feature maps into small groups.
      \textit{BN} refers to batch normalization~\cite{BatchNorm}.
      Shortcut connections of the architectures are summation except for those of DenseNet, which are concatenation.
    }\label{fig:block}
  \end{figure*}
  \begin{table*}[t]
    \centering
    \captionsetup{width=\linewidth}
    \caption{
      Network Architectures.
      Each convolutional layer is followed by batch normalization~\cite{BatchNorm} and a ReLU~\cite{ReLU}.
      Spatio-temporal down-sampling is performed by conv3\_1, conv4\_1, and conv5\_1 with a stride of two, except for DenseNet.
      \(F\) is the number of feature channels corresponding in Figure~\ref{fig:block}, and \(N\) is the number of blocks in each layer.
      DenseNet down-samples inputs using the transition layer,
      that consists of a \(3 \times 3 \times 3\) convolutional layer and a \(2 \times 2 \times 2\) average pooling layer with a stride of two,
      after conv2\_x, conv3\_x, and conv4\_x.
      \(F\) of DenseNet is the number of input feature channels of first block in each layer,
      and \(N\) is the same as that of the other networks.
      A \(3 \times 3 \times 3\) max-pooling layer (stride 2) is also located before conv2\_x of all networks for down-sampling.
      In addition, conv1 spatially down-samples inputs with a spatial stride of two.
      \(C\) of the fully-connected layer is the number of classes.
    }\label{tbl:network}
    \resizebox{\linewidth}{!}{
      \begin{tabular}{L{20mm}lcccccccccc}
        \toprule
        \multirow{2}{*}{Model} & \multirow{2}{*}{Block} &
          \multirow{2}{*}{conv1} &\multicolumn{2}{c}{conv2\_x} &
          \multicolumn{2}{c}{conv3\_x} & \multicolumn{2}{c}{conv4\_x} &
          \multicolumn{2}{c}{conv5\_x} & \multirow{2}{*}{} \\ \cmidrule(l){4-11}
        & & & \(F\) & \(N\) & \(F\) & \(N\) & \(F\) & \(N\) & \(F\) & \(N\) & \\ \midrule
        ResNet- \hspace{5mm} \{18, 34\} & Basic &
          \multirow{10}{*}{\rotatebox{90}{\parbox[c]{26mm}{
          \centering conv, \(7 \times 7 \times 7, 64\), \\ temporal stride 1, spatial stride 2}}}
          & 64 & \{2, 3\} & 128 & \{2, 4\} & 256 & \{2, 6\} & 512 & \{2, 3\} &
          \multirow{10}{*}{\rotatebox{90}{\parbox[c]{35mm}{
          \centering global average pool, \\
          \(C\)-d fully-connected, \\ softmax}}} \\
        \parbox[c]{28mm}{ResNet-\{50, \\ 101, 152, 200\}} & Bottleneck & &
          64 & 3 & 128 & \twotwo{4}{4}{8}{24} & 256 & \twotwo{6}{23}{36}{36} & 512 & 3 & \\
        Pre-act \hspace{5mm} ResNet-200 & Pre-act & & 64 & 3 & 128 & 24 & 256 & 36 & 512 & 3 & \\
        WRN-50 & Bottleneck & & 128 & 3 & 256 & 4 & 512 & 6 & 1024 & 3 & \\
        ResNeXt-101 & ResNeXt & & 128 & 3 & 256 & 24 & 512 & 36 & 1024 & 3 & \\
        DenseNet- \hspace{5mm} \{121, 201\} & DenseNet & & 64 & \{6, 6\} &
          128 & \{12, 12\} & 256 & \{24, 48\} & \oneone{512}{896} & \{16, 32\} & \\
        \bottomrule
      \end{tabular}
    }
  \end{table*}

  \subsection{Summary}
    In this study, in order to determine whether current video datasets have sufficient data for training of deep 3D CNNs,
    we conducted the three experiments described below
    using UCF-101~\cite{UCF101}, HMDB-51~\cite{HMDB51}, ActivityNet~\cite{activitynet}, and Kinetics~\cite{Kinetics}.
    We first examined the training of relatively shallow 3D CNNs from scratch on each video dataset.
    According to previous works~\cite{Hara_2017_ICCV_Workshops,Kinetics},
    3D CNNs trained on UCF-101, HMDB-51, and ActivityNet do not achieve high accuracy
    whereas ones trained on Kinetics work well.
    We try to reproduce such results to ascertain whether the datasets have sufficient data for deep 3D CNNs.
    Specifically, we used ResNet-18, which is the shallowest ResNet architecture,
    based on the assumption that if the ResNet-18 overfits when being trained on a dataset,
    that dataset is too small to be used for training deep 3D CNNs from scratch.
    See Section~\ref{sec:exp_resnet18} for details.

    We then conducted a separate experiment to determine
    whether the Kinetics dataset could train deeper 3D CNNs.
    A main point of this trial was to determine how deeply the datasets could train 3D CNNs.
    Therefore, we trained 3D ResNets on Kinetics while varying the model depth from 18 to 200.
    If Kinetics can train very deep CNNs, such as ResNet-152, which achieved the best performance in ResNets on ImageNet~\cite{ResNet},
    we can be confident that they have sufficient data to train 3D CNNs.
    Therefore, the results of this experiment are expected to be very important for the future progress
    in action recognition and other video tasks.
    See Section~\ref{sec:exp_kinetics} for details.

    In the final experiment, we examined the fine-tuning of Kinetics pretrained 3D CNNs on UCF-101 and HMDB-51.
    Since pretraining on large-scale datasets is an effective way to achieve good performance levels on small datasets,
    we expect that the deep 3D ResNets pretrained on Kinetics would perform well on relatively small UCF-101 and HMDB-51.
    This experiment examines 
    whether the transfer visual representations by deep 3D CNNs from one domain to another domain works effectively.
    See Section~\ref{sec:exp_ft} for details.

  \subsection{Network architectures}
    Next, we explain the various ResNet-based architectures with 3D convolutions used in this study.
    ResNet, which is one of the most successful architectures in image classification,
    provides shortcut connections that allow a signal to bypass one layer and move to the next layer in the sequence.
    Since these connections pass through the networks' gradient flows from the later layers to the early layers,
    they can facilitate the training of very deep networks.
    Unlike previous studies that examined only limited 3D ResNet architectures~\cite{Hara_2017_ICCV_Workshops,res3d},
    we examine not only deeper architectures but also some extended versions of ResNet.
    In particular, we explore the following architectures:
    ResNet (basic and bottleneck blocks)~\cite{ResNet},
    pre-activation ResNet~\cite{He2016}, wide ResNet (WRN)~\cite{WideResNet},
    ResNeXt~\cite{resnext}, and DenseNet~\cite{densenets}.
    The architectures are summarized in Figure~\ref{fig:block} and Table~\ref{tbl:network}.
    In the following paragraphs, we will briefly introduce each architecture.

    A basic ResNets block consists of two convolutional layers, and
    each convolutional layer is followed by batch normalization and a ReLU.\@
    A shortcut pass connects the top of the block to the layer just before the last ReLU in the block.
    ResNet-18 and 34 adopt the basic blocks.
    We use identity connections and zero padding for the shortcuts of the basic blocks (type A in~\cite{ResNet})
    to avoid increasing the number of parameters of these relatively shallow networks.

    A ResNets bottleneck block consists of three convolutional layers.
    The kernel sizes of the first and third convolutional layers are \(1 \times 1 \times 1\),
    whereas those of the second are \(3 \times 3 \times 3\).
    The shortcut pass of this block is the same as that of the basic block.
    ResNet-50, 101, 152, and 200 adopt the bottleneck.
    We use identity connections except for those that are used for increasing dimensions (type B in~\cite{ResNet}).

    The pre-activation ResNet is similar to bottleneck ResNet architectures,
    but there are differences in the convolution, batch normalization, and ReLU order.
    In ResNet, each convolutional layer is followed by batch normalization and a ReLU,
    whereas each batch normalization of the pre-activation ResNet is followed by the ReLU and a convolutional layer.
    A shortcut pass connects the top of the block to the layer just after the last convolutional layer in the block.
    In their study, He et al.\ showed that such pre-activation facilitates optimization in the training and reduces overfitting~\cite{He2016}.
    In this study, pre-activation ResNet-200 was evaluated.

    The WRN architecture is the same as the ResNet (bottleneck),
    but there are differences in the number of feature maps for each convolutional layer.
    WRN increases the number of feature maps rather than the number of layers.
    Such wide architectures are efficient in parallel computing using GPUs~\cite{WideResNet}.
    In this study, we evaluate the WRN-50 using a widening factor of two.
    
    ResNeXt introduces cardinality, which is a different dimension from deeper and wider.
    Unlike the original bottleneck block, the ResNeXt block introduces group convolutions,
    which divide the feature maps into small groups.
    Cardinality refers to the number of middle convolutional layer groups in the bottleneck block.
    In their study, Xie et al.\ showed that
    increasing the cardinality of 2D architectures is more effective than using wider or deeper ones~\cite{resnext}.
    In this study, we evaluate ResNeXt-101 using the cardinality of 32.

    DenseNet makes connections from early layers to later layers by
    the use of a concatenation that is different from the ResNets summation.
    This concatenation connects each layer densely in a feed-forward fashion.
    DenseNets also adopt the pre-activation used in pre-activation ResNets.
    In their study, Huang et al.\ showed that it achieves better accuracy with fewer parameters than ResNets~\cite{densenets}.
    In this study, we evaluate DenseNet-121 and 201 using a growth rate of 32.

  \subsection{Implementation}
    \noindent \textbf{Training.}
      We use stochastic gradient descent with momentum to train the networks and
      randomly generate training samples from videos in training data in order to perform data augmentation.
      First, we select a temporal position in a video by uniform sampling in order to generate a training sample.
      A 16-frame clip is then generated around the selected temporal position.
      If the video is shorter than 16 frames, then we loop it as many times as necessary.
      Next, we randomly select a spatial position from the 4 corners or the center.
      In addition to the spatial position, we also select a spatial scale of the sample in order to perform multi-scale cropping.
      The procedure used is the same as~\cite{VeryDeepTwo}.
      The scale is selected from \(\left \{1, \frac{1}{2^{1/4}}, \frac{1}{\sqrt{2}}, \frac{1}{2^{3/4}}, \frac{1}{2}\right \} \).
      Scale 1 means that the sample width and height are the same as the short side length of the frame,
      and scale 0.5 means that the sample is half the size of the short side length.
      The sample aspect ratio is 1 and the sample is spatio-temporally cropped at the positions, scale, and aspect ratio.
      We spatially resize the sample at \(112 \times 112\) pixels.
      The size of each sample is
      \(3\) channels \(\times \ 16\) frames \(\times \ 112\) pixels \(\times \ 112\) pixels,
      and each sample is horizontally flipped with \( 50\% \) probability.
      We also perform mean subtraction,
      which means that we subtract the mean values of ActivityNet from the sample for each color channel.
      All generated samples retain the same class labels as their original videos.

      In our training, we use cross-entropy losses and back-propagate their gradients.
      The training parameters include a weight decay of 0.001 and 0.9 for momentum.
      When training the networks from scratch, we start from learning rate 0.1,
      and divide it by 10 after the validation loss saturates.
      When performing fine tuning, we start from a learning rate of 0.001, and assign a weight decay of 1e-5.

    \vspace{1mm}
    \noindent \textbf{Recognition.}
      We adopt the sliding window manner to generate input clips,
      (i.e., each video is split into non-overlapped 16-frame clips),
      and recognize actions in videos using the trained networks.
      Each clip is spatially cropped around a center position at scale 1.
      We then input each clip into the networks and
      estimate the clip class scores, which are averaged over all the clips of the video.
      The class that has the maximum score indicates the recognized class label.

  \subsection{Datasets}
    \begin{figure*}[t]
      \begin{subfigure}[b]{0.2475\linewidth}
        \includegraphics[width=\linewidth]{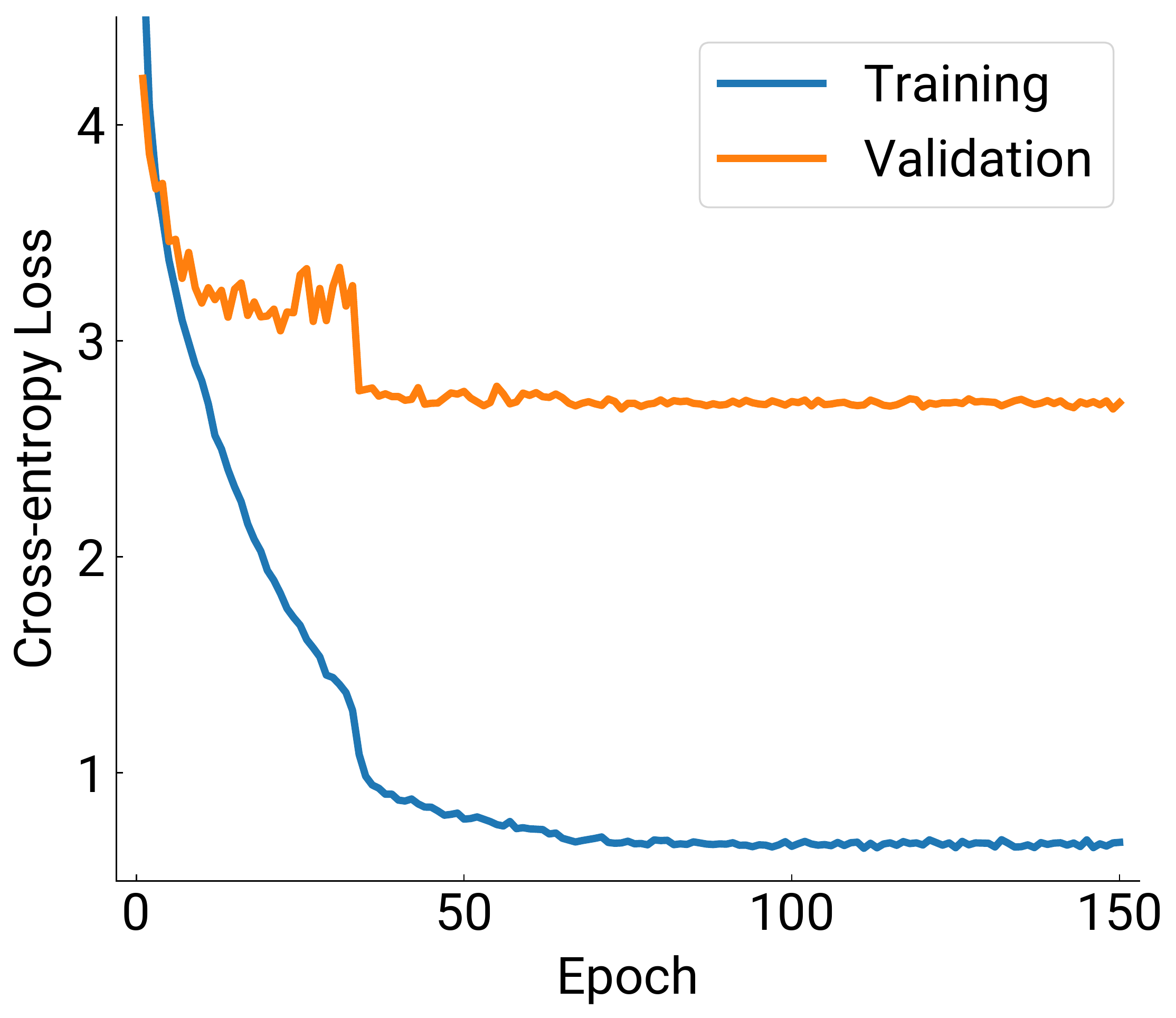}
        \caption{UCF-101 (split 1).}\label{fig:resnet18_ucf101}
      \end{subfigure}
      \begin{subfigure}[b]{0.2475\linewidth}
        \includegraphics[width=\linewidth]{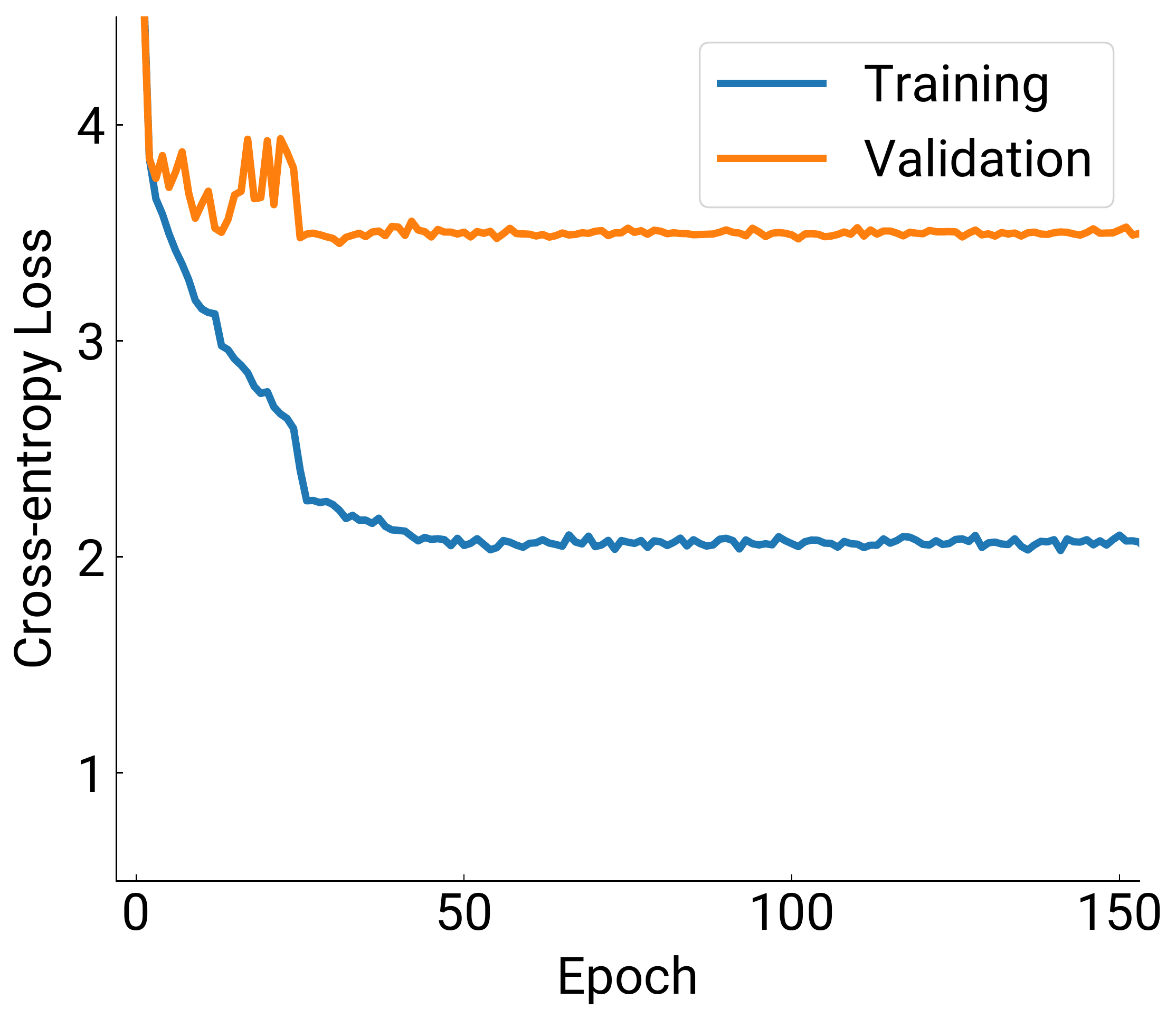}
        \caption{HMDB-51 (split 1).}\label{fig:resnet18_hmdb51}
      \end{subfigure}
      \begin{subfigure}[b]{0.2475\linewidth}
        \includegraphics[width=\linewidth]{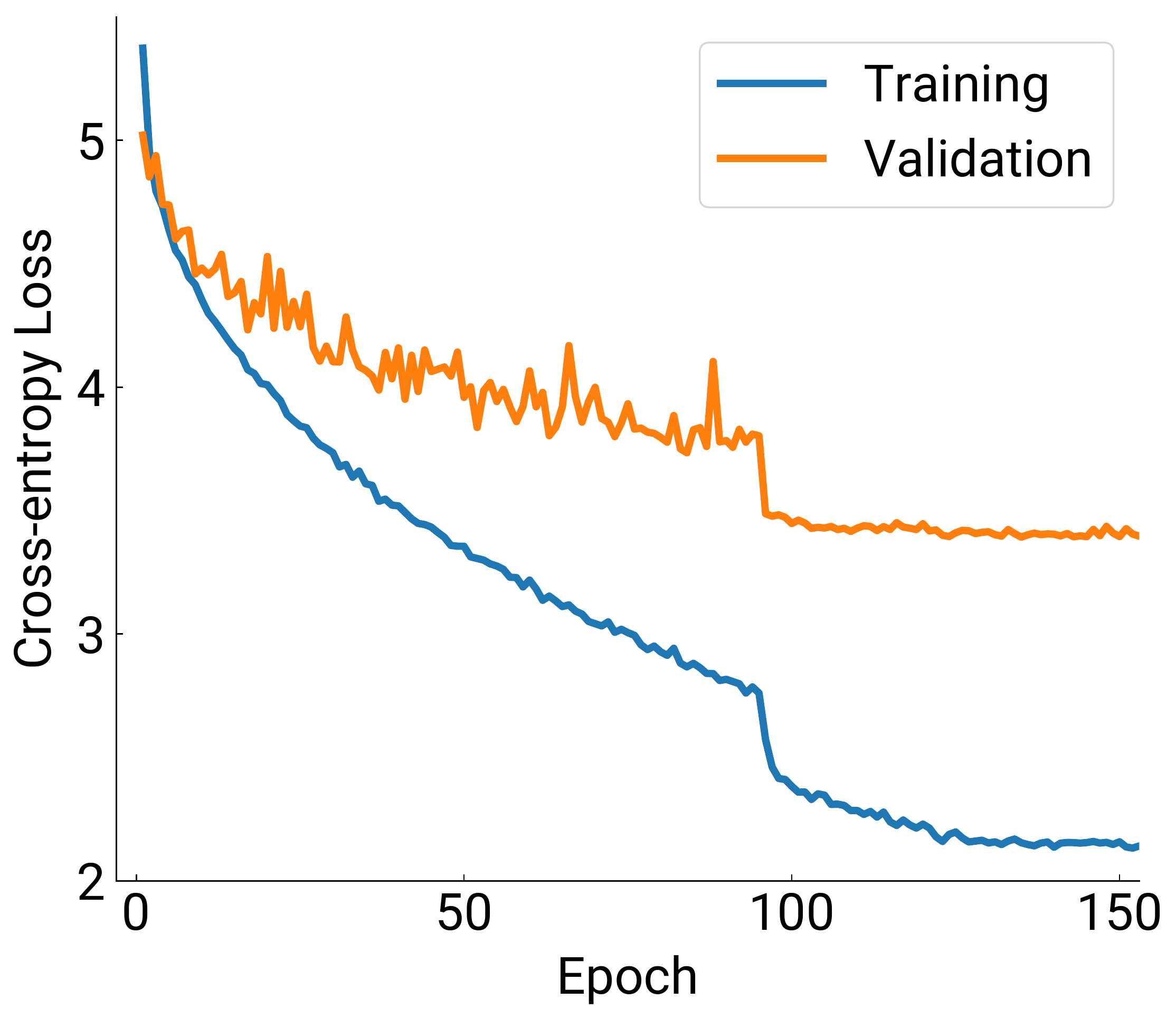}
        \caption{ActivityNet.}\label{fig:resnet18_activitynet}
      \end{subfigure}
      \begin{subfigure}[b]{0.2475\linewidth}
        \includegraphics[width=\linewidth]{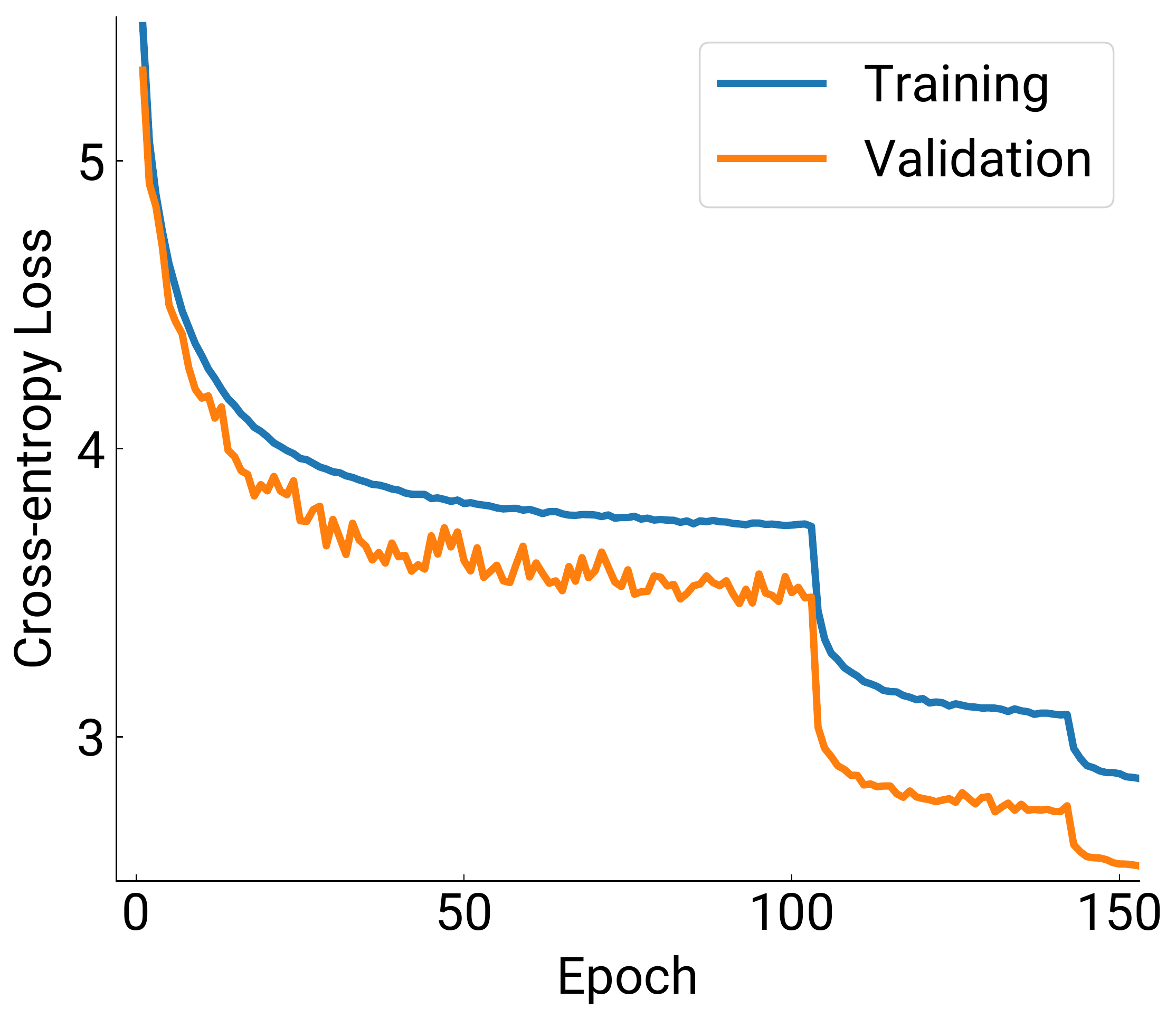}
        \caption{Kinetics.}\label{fig:resnet18_kinetics}
      \end{subfigure}
      \caption{
        ResNet-18 training and validation losses.
        The validation losses on UCF-101, HMDB-51, and ActivityNet quickly converged to high values
        and were clearly higher than their corresponding training losses.
        The validation losses on Kinetics were slightly higher than the corresponding training losses,
        significantly different than those on the other datasets.
      }\label{fig:resnet18}
    \end{figure*}
    As stated above, this study focuses on four datasets:
    UCF-101~\cite{UCF101}, HMDB-51~\cite{HMDB51}, ActivityNet~\cite{activitynet}, and Kinetics~\cite{Kinetics}.

    UCF-101 includes 13,320 action instances from 101 human action classes.
    The videos were temporally trimmed to remove non-action frames.
    The average duration of each video is about 7 seconds.
    Three train/test splits (70\% training and 30\% testing) are provided in the dataset.

    HMDB-51 includes 6,766 videos from 51 human action classes.
    Similar to UCF-101, the videos were temporally trimmed.
    The average duration of each video is about 3 seconds.
    Three train/test splits (70\% training and 30\% testing) are provided in this dataset.

    ActivityNet (v1.3) provides samples from 200 human action classes with an average of 137 untrimmed videos per class and 1.41 activity instances per video.
    Unlike the other datasets, ActivityNet consists of untrimmed videos, which include non-action frames.
    The total video length is 849 hours,
    and the total number of action instances is 28,108.
    This dataset is randomly split into three different subsets:
    training, validation, and testing.
    More specifically, 50\% is used for training, 25\% is used for validation, and 25\% is used for testing.

    The Kinetics dataset has 400 human action classes,
    and consists of more than 400 videos for each class.
    The videos were temporally trimmed and last around 10 seconds.
    The total number of the videos is in excess of 300,000.
    The number of training, validation, and testing sets are about
    240,000, 20,000, and 40,000, respectively.

    The video properties of all these datasets are similar.
    Most videos were extracted from YouTube, except for HMDB-51,
    which includes videos extracted from movies.
    The videos include dynamic background and camera motions and
    the main differences among them are the numbers of action classes and instances.

    We resized the videos to heights of 240 pixels without changing their aspect ratios and then stored them.

\section{Results and discussion}\label{sec:exp}
  \subsection{Analyses of training on each dataset}\label{sec:exp_resnet18}
    We began by training ResNet-18 on each dataset.
    According to previous works~\cite{Hara_2017_ICCV_Workshops,Kinetics},
    3D CNNs trained on UCF-101, HMDB-51, and ActivityNet do not achieve high accuracy
    whereas ones trained on Kinetics work well.
    We tried to reproduce such results in this experiment.
    In this process, we used split 1 of UCF-101 and HMDB-51,
    and the training and validation sets of ActivityNet and Kinetics.

    Figure~\ref{fig:resnet18} shows the training and validation losses of ResNet-18 on each dataset.
    As can be seen in the figure, the validation losses on UCF-101, HMDB-51, and ActivityNet quickly converged to high values
    and were clearly higher than their corresponding training losses.
    These results indicate that overfitting resulted when the training on those three datasets.
    In addition to those losses, we confirmed per-clip accuracies,
    which are evaluated for each clip rather than for each video.
    The validation accuracies of UCF-101, HMDB-51, and ActivityNet are 40.1, 16.2, and 26.8\%, respectively.
    It should be noted that direct comparisons between our results and those of previous studies would be unfair
    because the accuracies reported in most papers were per-video accuracies.
    However, since these accuracies are very low even compared with earlier methods~\cite{activitynet,Wang2013},
    our results indicate that it is difficult to train deep 3D CNNs from scratch on UCF-101, HMDB-51, and ActivityNet.

    In contrast, the training and validation losses on Kinetics are significantly different than those on other datasets.
    Since the validation losses were slightly higher than the training losses,
    we could conclude that training ResNet-18 on Kinetics did not result in overfitting,
    and that it is possible for Kinetics to train deep 3D CNNs.
    In the next section, we will further investigate deeper 3D CNNs on Kinetics.

  \subsection{Analyses of deeper networks}\label{sec:exp_kinetics}
    Since the abovementioned experiment showed Kinetics could be used to train ResNet-18 without overfitting,
    we next examined deeper ResNets using the Kinetics training and validation sets.
    
    Here, we will show ResNets accuracies changes based on model depths.
    Figure~\ref{fig:resnet_depth} shows the averaged accuracies over top-1 and top-5 ones.
    We can see that, essentially, as the depth increased, accuracies improved,
    and that the improvements continued until reaching the depth of 152.
    We can also see that deeper ResNet-152 achieved significant improvement of accuracies
    compared with ResNet-18, which was the previously examined architecture~\cite{Hara_2017_ICCV_Workshops,res3d}.
    %
    In contrast, the accuracy of ResNet-200 was almost the same as that of ResNet-152.
    This result indicate that the training of ResNet-200 started to overfit.
    Interestingly, the results are similar to those for 2D ResNets on ImageNet~\cite{He2016}.
    More specifically, the accuracies of both 2D and 3D ResNets improved as the depth increased until reaching the depth of 152,
    and then the accuracies did not increase when increasing the depths of 200.
    These results indicate that the Kinetics dataset has sufficient data to train 3D CNNs in a manner similar to ImageNet.

    Comparisons with other architectures are shown in Table~\ref{tbl:kinetics_val}.
    Here, it can be seen that the accuracies of pre-activation ResNet-200 are slightly low when compared with the standard ResNet-200
    though He et al.\ reported that the pre-activation reduces overfitting and improves 2D ResNet-200 on ImageNet~\cite{He2016}.
    We can also see that the WRN-50 achieved higher accuracies when compared with the ResNet-152,
    which is similar to the results on ImageNet~\cite{WideResNet}.
    This result also supports that Kinetics is sufficient large for the training of 3D CNNs
    because the number of parameters of WRN-50 is larger than the ResNet-152.
    Furthermore, we can see that ResNeXt-101 achieved the best accuracies among the architectures tested.
    This result is also similar to that seen for ImageNet~\cite{resnext},
    and means that introducing the cardinality works well for the 3D ResNets on Kinetics.
    In contrast, the accuracies of the DenseNet-121 and 201 were slightly lower than the other architectures.
    The major advantage provide by dense connections is parameter efficiency,
    which contributes to reducing overfitting~\cite{densenets}.
    However, Kinetics did not need such techniques to train deep 3D CNNs.

    Table~\ref{tbl:kinetics_test} shows the results of the Kinetics test set
    used to compare ResNeXt-101, which achieved the highest accuracies, with the state-of-the-art methods.
    Here, it can be seen that the accuracies of ResNeXt-101 are clearly high
    compared with C3D with batch normalization~\cite{Kinetics}, which is 10-layer network,
    as well as CNN+LSTM and two-stream CNN~\cite{Kinetics}.
    This result also indicates the effectiveness of deeper 3D networks trained on Kinetics.
    In contrast, RGB-I3D trained on Kinetics from scratch~\cite{I3D_arxiv} were found to outperform ResNeXt-101
    even though ResNeXt-101 is a deeper architecture than I3D.
    One of the reasons for this is the size differences of the network inputs.
    Specifically, the size of I3D is \(3 \times 64 \times 224 \times 224\),
    whereas that of ResNeXt-101 is \(3 \times 16 \times 112 \times 112\).
    Thus, I3D is 64 times larger than ResNeXt-101.
    To confirm the accuracies when using larger inputs,
    we also evaluated the ResNeXt-101 that used \(3 \times 64 \times 112 \times 112\) inputs,
    which are the largest available sizes in our environment (NVIDIA TITAN X \(\times \ 8\)).
    We can see that the network, referred as ResNeXt-101 (64f) in Table~\ref{tbl:kinetics_test},
    outperformed RGB-I3D even though the input size is still four times smaller than that of I3D.
    We can conclude that deeper 3D architectures trained on Kinetics are effective.
    In addition, it is felt that combining two-stream architectures with ResNeXt-101 make further improvements
    based on higher accuracies of two-stream I3D.
    
    \begin{table}[t]
      \centering
      \captionsetup{width=0.95\linewidth}
      \caption{
        Accuracies on the Kinetics validation set.
        \textit{Average} is averaged accuracy over \textit{Top-1} and \textit{Top-5}.
      }\label{tbl:kinetics_val}
      \begin{tabular}{lccc}
        \toprule
        Method & Top-1 & Top-5 & Average \\
        \midrule
        ResNet-18 & 54.2 & 78.1 & 66.1 \\
        ResNet-34 & 60.1 & 81.9 & 71.0 \\
        ResNet-50 & 61.3 & 83.1 & 72.2 \\
        ResNet-101 & 62.8 & 83.9 & 73.3 \\
        ResNet-152 & 63.0 & \textbf{84.4} & \textbf{73.7} \\
        ResNet-200 & \textbf{63.1} & \textbf{84.4} & \textbf{73.7} \\
        \midrule
        ResNet-200 (pre-act) & 63.0 & 83.7 & 73.4 \\
        Wide ResNet-50 & 64.1 & 85.3 & 74.7 \\
        ResNeXt-101 & \textbf{65.1} & \textbf{85.7} & \textbf{75.4} \\
        DenseNet-121 & 59.7 & 81.9 & 70.8 \\
        DenseNet-201 & 61.3 & 83.3 & 72.3 \\
        \bottomrule          
      \end{tabular}
    \end{table}
    \begin{table}[t]
      \centering
      \caption{
        Accuracies on the Kinetics test set.
        \textit{Average} is averaged accuracy over \textit{Top-1} and \textit{Top-5}.
        Here, we refer the results of RGB- and Two-stream I3D trained from scratch~\cite{I3D_arxiv} for fair comparison.
      }\label{tbl:kinetics_test}
      \begin{tabular}{L{37.8mm}ccc}
        \toprule
        Method & Top-1 & Top-5 & Average \\
        \midrule
        ResNeXt-101 & -- & -- & 74.5 \\
        ResNeXt-101 (64f) & -- & -- & \textbf{78.4} \\
        \midrule
        CNN+LSTM~\cite{Kinetics} & 57.0 & 79.0 & 68.0 \\
        Two-stream CNN~\cite{Kinetics} & 61.0 & 81.3 & 71.2 \\
        C3D w/ BN~\cite{Kinetics} & 56.1 & 79.5 & 67.8 \\
        RGB-I3D~\cite{I3D_arxiv} & 68.4 & 88.0 & 78.2 \\
        Two-stream I3D~\cite{I3D_arxiv} & 71.6 & 90.0 & \textbf{80.8} \\
        \bottomrule
      \end{tabular}
    \end{table}

  \subsection{Analyses of fine-tuning}\label{sec:exp_ft}
    Finally, in this section we confirm the performance of fine-tuning.
    In the experiments above, we showed that Kinetics can train deep 3D CNNs from scratch,
    but that it is difficult to train such networks on other datasets.
    In this section, we fine-tuned the Kinetics pretrained 3D CNNs on UCF-101 and HMDB-51.
    The results of this experiment are important for determining whether the 3D CNNs are effective for other datasets.
    It should be noted that, in this experiment, fine-tuning was only performed to train conv5\_x and the fully connected layer
    because it achieved the best performance during the preliminary experiments.

    \begin{table}[t]
      \centering
      \caption{
        Top-1 accuracies on UCF-101 and HMDB-51.
        All accuracies are averaged over three splits.
      }\label{tbl:ucf101_hmdb51}
      \begin{tabular}{lcc}
        \toprule
        Method & UCF-101 & HMDB-51 \\
        \midrule
        ResNet-18 (scratch) & 42.4 & 17.1 \\
        \midrule
        ResNet-18 & 84.4 & 56.4 \\
        ResNet-34 & 87.7 & 59.1 \\
        ResNet-50 & 89.3 & 61.0 \\
        ResNet-101 & 88.9 & 61.7 \\
        ResNet-152 & 89.6 & 62.4 \\
        ResNet-200 & 89.6 & 63.5 \\
        \midrule
        DenseNet-121 & 87.6 & 59.6 \\
        ResNeXt-101 & \textbf{90.7} & \textbf{63.8} \\
        \bottomrule
      \end{tabular}
    \end{table}
    Table~\ref{tbl:ucf101_hmdb51} shows the accuracies of Kinetics pretrained 3D CNNs,
    as well as ResNet-18 trained from scratch, in UCF-101 and HMDB-51.
    Here, it can be seen that Kinetics pretrained ResNet-18 clearly outperformed one trained from scratch.
    This result indicate that pretraining on Kinetics is effective on UCF-101 and HMDB-51.
    We can also see that the accuracies basically improved as the depth increased,
    similar to the results obtained on Kinetics.
    However, unlike the results on Kinetics, ResNet-200 also improved the accuracies in HMDB-51.
    Because, as described above, the fine-tuning in this experiment
    was only performed to train conv5\_x and the fully connected layer,
    the numbers of trained parameters were the same from ResNet-50 to ResNet-200.
    Therefore, the pretrained early layers, which work as feature extractors,
    relate to the differences of performance.
    These results indicate that feature representations of ResNet-200 would be suitable for HMDB-51
    even though the 200-layer network might start to overfit on Kinetics.

    Table~\ref{tbl:ucf101_hmdb51} also shows that ResNeXt-101, which achieved the best performance on Kinetics,
    achieved the highest levels of performance on both datasets when compared with the other networks.
    The performance difference, however, is smaller than that of Kinetics.
    It is considered likely that this result also relates to the sizes of datasets.
    We then compared the results with DenseNet-121
    because it is a parameter-efficient network,
    and thus might achieve better performance on small datasets.
    However, the DenseNet-121 results were lower than those of ResNet-50,
    thereby indicating that its greater efficiency did not contribute on fine-tuning of 3D CNNs.

    We shows the results of our comparison with state-of-the-art methods in Table~\ref{tbl:ucf101_hmdb51_sota}.
    Here, we can see that ResNeXt-101 achieved higher accuracies compared with
    C3D~\cite{C3D}, P3D~\cite{Qiu_2017_ICCV}, two-stream CNN~\cite{Simonyan2014}, and TDD~\cite{Wang2015TDD}.
    Furthermore, we can also see that ResNeXt-101 (64f), which utilize larger inputs described in previous section,
    slightly outperformed ST Multiplier Net~\cite{feichtenhofer2017multiplier} and TSN~\cite{Wang2016TSN},
    which utilize more complex two-stream architectures.
    We can also see that two-stream I3D~\cite{I3D_arxiv}, which utilizes simple two-stream 3D architectures pretrained on Kinetics,
    achieved the best accuracies.
    Based on these results, we can conclude that simple 3D architectures pretrained on Kinetics outperform complex 2D architectures.
    We believe that development of 3D CNNs rapidly grows and contributes to significant advances in video recognition and its related tasks.
    \begin{table}[t]
      \centering
      \caption{
        Top-1 accuracies on UCF-101 and HMDB-51 comared with the state-of-the-art methods.
        All accuracies are averaged over three splits.
        \textit{Dim} indicate the dimension of convolution kernel.
      }\label{tbl:ucf101_hmdb51_sota}
      \begin{tabular}{lccc}
        \toprule
        Method & Dim & UCF-101 & HMDB-51 \\
        \midrule
        ResNeXt-101 & 3D & 90.7 & 63.8 \\
        ResNeXt-101 (64f) & 3D & \textbf{94.5} & \textbf{70.2} \\
        \midrule
        C3D~\cite{C3D} & 3D & 82.3 & -- \\
        P3D~\cite{Qiu_2017_ICCV} & 3D & 88.6 & -- \\
        Two-stream I3D~\cite{I3D_arxiv} & 3D & \textbf{98.0} & \textbf{80.7} \\
        \midrule
        Two-stream CNN~\cite{Simonyan2014} & 2D & 88.0 & 59.4 \\
        TDD~\cite{Wang2015TDD} & 2D & 90.3 & 63.2 \\
        ST Multiplier Net~\cite{feichtenhofer2017multiplier} & 2D & 94.2 & 68.9 \\
        TSN~\cite{Wang2016TSN} & 2D & 94.2 & 69.4 \\
        \bottomrule
      \end{tabular}
    \end{table}

\section{Conclusion}
  In this study, we examined the architectures of various CNNs with spatio-temporal 3D convolutional kernels on current video datasets.
  Based on the results of those experiments, the following conclusions could be obtained:
  (i) ResNet-18 training resulted in significant overfitting for UCF-101, HMDB-51, and ActivityNet but not for Kinetics.
  (ii) The Kinetics dataset has sufficient data for training of deep 3D CNNs,
  and enables training of up to 152 ResNets layers, interestingly similar to 2D ResNets on ImageNet.
  (iii) Kinetics pretrained simple 3D architectures outperforms complex 2D architectures on UCF-101 and HMDB-51,
  and the pretrained ResNeXt-101 achieved 94.5\% and 70.2\% on UCF-101 and HMDB-51, respectively.

  We believe that the results of this study will facilitate further advances in video recognition and its related tasks.
  Following the significant advances in image recognition made by 2D CNNs and ImageNet,
  pretrained 2D CNNs on ImageNet experienced significant progress in various tasks
  such as object detection, semantic segmentation, and image captioning.
  It is felt that, similar to these,
  3D CNNs and Kinetics have the potential to contribute to significant progress in fields related to various video tasks
  such as action detection, video summarization, and optical flow estimation.
  In our future work, we will investigate transfer learning not only for action recognition but also for other such tasks.

{\small
\bibliographystyle{ieee}
  
}

\end{document}